\pdfoutput=1

\documentclass[11pt]{article}

\usepackage[final]{acl}

\usepackage{times}
\usepackage{latexsym}

\usepackage{algorithm}
\usepackage{algpseudocode}
\usepackage{multirow,adjustbox,makecell}
\usepackage{booktabs}
\usepackage{amssymb}
\usepackage{adjustbox}
\usepackage{subcaption}
\usepackage{calc} 

\usepackage[T1]{fontenc}

\usepackage[utf8]{inputenc}

\usepackage{microtype}

\usepackage{inconsolata}

\usepackage{graphicx}
\usepackage{amsmath}

\title{Mitigating Attention Localization in Small Scale: \\
Self-Attention Refinement via One-step Belief Propagation}

\author{
 \textbf{Nakyung Lee\textsuperscript{1}},
 \textbf{Yeongoon Kim\textsuperscript{1}},
 \textbf{Minhae Oh\textsuperscript{1}},
 \textbf{Suhwan Kim\textsuperscript{1}},
 \textbf{Jin Woo Koo\textsuperscript{1}},
 \textbf{Hyewon Jo\textsuperscript{1}},
 \textbf{Jungwoo Lee\textsuperscript{1}}
\\
 \textsuperscript{1}Seoul National University \\
 {\small
   \href{mailto:leena@cml.snu.ac.kr}{leena@cml.snu.ac.kr} (first author), 
   \href{mailto:junglee@snu.ac.kr}{junglee@snu.ac.kr} (corresponding author)}
}

\begin{document}
\maketitle
\begin{abstract}
Transformer-based self-attention mechanism serves as the core of modern language models, yet it often suffers from \textit{localization}, where attentions collapse onto a limited subset of tokens and fail to capture long-range dependencies. To address this issue, we propose \textbf{Self-Attention One-step Belief Propagation (SAOBP)}, a refinement framework that injects \emph{multi-hop} relationships through a belief propagation process\footnote{The code is released at \href{https://github.com/nakyungLee20/SAOBP.git}{nakyungLee20/SAOBP}}. To interpret and quantify these interactions, we introduce \textbf{Global Token Dependency (GTD)} that captures the relative contribution of \emph{multi-hop} connections within the attention graph. Empirical results indicate that SAOBP helps prevent entropy collapse in deeper layers and adaptively maintains GTD at task-appropriate levels, thereby supporting improvements in model performance. Importantly, we observe competitive gains in small-scale models, highlighting its potential for improving inference quality in resource-constrained scenarios.
\end{abstract}

\section{Introduction}
Transformer architecture~\cite{Vaswani2017trans} forms the backbone of modern large language models (LLMs), demonstrating remarkable performance across diverse natural language processing tasks. At the core of these models lies the self-attention mechanism, which enables dynamic modeling of contextual relationships between tokens. 

Despite its empirical success, several studies have identified limitations associated with attention \textit{localization}, a phenomenon in which attention distributions collapse onto a limited subset of tokens. Such localized attention tends to have low entropy~\cite{dong2024transformer}, low-rank sparsity~\cite{bao2024eigen} and sparsed attention (Fig.~\ref{app_fig:sparse-attn} in the appendix), which negatively affect the representational power, training stability, and downstream model performance~\cite{zhai2023stabilizing}. This problem is particularly pronounced in small-scale Transformer variants. Their limited depth and width inherently restrict the ability to propagate information across layers, making them more susceptible to localized attention than larger models.

\begin{figure*}[ht]
\begin{center}
\centerline{\includegraphics[width=\textwidth]{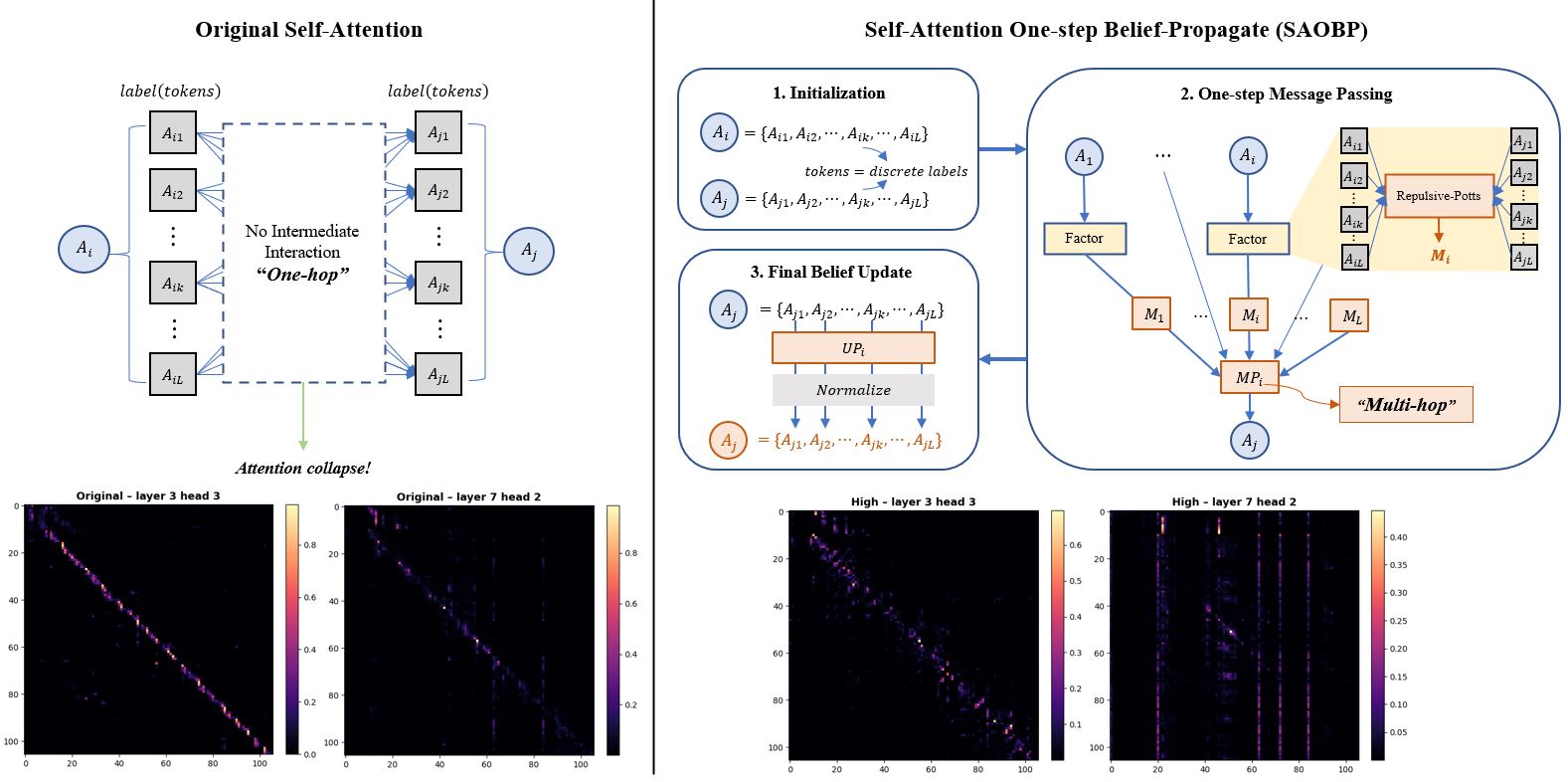}}
\caption {Comparison between original self-attention (left) and proposed SAOBP algorithm (right): The standard self-attention formulation relies on direct (one-hop) token interactions, which often yield sparse attention distributions and entropy collapse, as illustrated in the heatmaps. In contrast, SAOBP utilizes a structured message-passing step that propagates multi-hop dependency signals via belief updates governed by a repulsive Potts prior. This enhances global context modeling and redistributes attention mass more broadly, mitigating localization and improving representational diversity in deeper layers.}
\label{fig:saobp-framework}
\end{center}
\vskip -0.3in
\end{figure*}

In standard self-attention mechanisms, each query token primarily attends to a few highly salient keys~\cite{shisparse2021}, thus modeling predominantly \textit{one-hop\footnote{In this paper, we consider "local" as one-hop token dependencies, and "global" as multi-hop token dependencies.}} dependencies. To address this limitation, we hypothesize that explicitly incorporating \textit{global} context---defined here as multi-hop information flow---could mitigate the localization issue, particularly in compact architectures. To test this hypothesis, we propose \textbf{Self-Attention One-step Belief Propagation (SAOBP)}, a method that integrates global interactions by leveraging belief propagation (BP) principles combined with a repulsive Potts prior (Fig.~\ref{fig:saobp-framework}). Our empirical evaluations show that SAOBP effectively alleviates attention collapse and consistently enhances performance across downstream tasks.

To systematically investigate how SAOBP mitigates localization through multi-hop information flow and interpret its impact on model performance, we introduce \textbf{Global Token Dependency (GTD)}. GTD quantifies the relative attention mass contributed by intermediate, multi-hop transitions ($\geq 2$ hops) within the stochastic attention graph. Our empirical analysis illustrated in Fig.~\ref{fig:gtd-mp-corr} and Fig.~\ref{fig:gtd-analysis} demonstrates that GTD serves as a principled diagnostic tool, capable of detecting specific layers and heads where attention collapses into overly localized patterns.

\vspace{0.1in}
\textbf{Contributions}
\vspace{-0.1in}
\begin{itemize}
    \setlength\itemsep{-0.5em}
    \item We propose \textbf{SAOBP}, a self-attention regularization framework that suppresses entropy collapse and promotes diverse, globally-aware attention distributions.
    \item We introduce \textbf{Global Token Dependency (GTD)}, a novel diagnostic concept that quantifies the extent of multi-hop information flow in attention graphs. This value offers a principled measure of intermediate context modeling and attention localization.
    \item We empirically demonstrate that SAOBP improves model performance across a variety of downstream tasks. The benefits are more pronounced in small-scale settings, providing practical guidance for the design of efficient and expressive compact language models.
\end{itemize}
\vspace{-0.4em}

\section{Related works}
\paragraph{Small-scale models in NLP} 
Some recent studies have emphasized the practical advantages of small-scale language models, including cost-efficiency, rapid customization, and deployment feasibility in resource-constrained environments~\cite{wang2024comprehensive}. \citet{liu2024mobilellm} and \citet{thawakar2024mobillama} propose sub-billion parameter models optimized for mobile and edge applications. Similarly, \citet{jiao2020tinybert} and \citet{turc2019wellread} have demonstrated that small transformer models can achieve competitive performance through effective knowledge distillation and optimized architectures. Nonetheless, compact models typically underperform compared to larger-scale alternatives, and bridging this performance gap through targeted architectural enhancements remains an important open challenge. In this work, we introduce a framework that improves the expressiveness of compact models (with $\leq 50$M parameters), enhancing their utility in constrained settings.

\paragraph{Localization of Self-Attention}
Self-attention mechanisms often exhibit localization, where attention mass concentrates on a few tokens, leading to entropy collapse (where attention entropy skewed to zero) and representational degradation~\cite{Zhang2024AttentionEntropy, qi2025taming}. Prior works attribute this to low-rank eigen-spectra of query-key matrices~\cite{bao2024eigen} and the exponential nature of the softmax function~\cite{dong2024transformer}. While several regularization strategies have been proposed~\cite{jha2025entropy, zhai2023stabilizing}, these approaches do not explicitly quantify the extent of attention localization, making their effectiveness difficult to interpret. To overcome this limitation, we introduce GTD, a metric enabling interpretable, layer-wise analysis of attention spread.

\paragraph{Belief Propagation for Marginalization}
Belief propagation (BP) is a classical inference algorithm that enables efficient marginalization in structured graphical models~\cite{Pearl1988,Yedidia2001,Koller2009}. Recent studies have explored integrating BP into neural architectures for structured prediction~\cite{kim2017structured, Kuck2020BPNN, Dai2016Discriminative}. However, few studies have directly leveraged BP to refine attention patterns. Since each row of the attention matrix can be interpreted as a categorical probability distribution over tokens, BP can offer a theoretically grounded mechanism to reshape probability distributions under higher-order structural constraints. Our proposed SAOBP represents a distinct application of BP, utilizing its message-passing mechanism to inject multi-hop dependencies into attention and thereby mitigate localization effects.

\section{Preliminaries}
\paragraph{Self-Attention and Attention Entropy} Self-attention is mathematically defined by Eq.~\eqref{eq:attn}, where $Q$, $K$, and $V$ represent learnable weight matrices, and $B$, $H$, and $L$ denote the batch size, number of attention heads, and sequence length:
\begin{equation}
A = \text{softmax}\left(\frac{QK^T}{\sqrt{d}}\right) \quad \in \mathbb{R}^{B \times H \times L \times L}. \label{eq:attn}
\end{equation}
The internal dynamics of self-attention can be analyzed using the concept of attention entropy~\cite{Zhang2024AttentionEntropy}, defined formally in Eq.~\eqref{eq:attn_ent}. Attention entropy quantifies the uncertainty in attention distributions, offering insight into how information is structured within models. We leverage this metric to investigate attention entropy collapse.
\begin{equation}
\begin{split}
\mathcal{H}(A_i) &= -\sum_{j=1}^{L} A_{ij}\,\log A_{ij},\\
\mathcal{H}(A)   &= \frac{1}{L}\sum_{i=1}^{L} \mathcal{H}(A_i)
\end{split}
\label{eq:attn_ent}
\end{equation}

\paragraph{Constructing Belief Propagation} 
A belief propagation algorithm can be constructed with the following four components~\cite{BPtheory}. Let $i$ and $j$ be variable nodes connected by a factor node $f_{ij}$.

\textbf{1. Messages}: Messages represent probabilistic information passed between variable and factor nodes during the inference process. Formally, messages are denoted by $\tilde{m}_{f_{ij}\rightarrow i}, m_{i\rightarrow f_{ij}} \in \mathbb{M}$, where $\mathbb{M}$ is the message space. 

\textbf{2. Update functions (Factor functions)}: Message updates are governed by predefined functions, which determine the dynamics and convergence properties of the algorithm. Common variants of belief propagation include the sum-product and min-sum algorithms. Let $f_{ij}(r, k)$ denote the pairwise potential between labels $r$ and $k$ at the factor node connecting variables $i$ and $j$. At iteration $t$, the variable-to-factor and factor-to-variable messages are updated according to Eq.~\eqref{eq:msg-var2fac} and Eq.~\eqref{eq:msg-fac2var}: 
\begin{equation}
m^{(t+1)}_{i \to f_{ij}}(k) \;\propto\; \prod_{n \in \mathcal{N}(i)\setminus\{j\}} \tilde m^{(t)}_{f_{ni} \to i}(k) \label{eq:msg-var2fac}
\end{equation}
\begin{equation}
\tilde m^{(t)}_{f_{ij} \to j}(k) \;\propto\; \sum_{r} \psi_{ij}(r, k)\, m^{(t)}_{i \to f_{ij}}(r) \label{eq:msg-fac2var}
\end{equation}
where $\mathcal{N}(i)$ represents the set of factor nodes adjacent to variable node $i$.

\textbf{3. Initialization}: Messages and node states are initialized based on predefined criteria or heuristics, which affect the algorithm's dynamics and convergence. Initial messages may be uniform or informed by external evidences, such as priors $\phi_i(k)$ associated with node $i$. 

\textbf{4. Decision rule}: After a finite number of message-passing iterations, each node computes its belief over the label space by aggregating the incoming messages. This marginalization process is formally expressed as Eq.~\eqref{eq:final‐belief}, where $Z_i$ is a normalization constant and $\partial i$ denotes the set of factor nodes connected to node $i$.
\begin{equation}
b_i(x_i) = \frac{1}{Z_i} \prod_{f_{ji}\,\in\,\partial i} \tilde m_{\,f_{ji}\to i}\bigl(x_i\bigr) \label{eq:final‐belief}
\end{equation}

\paragraph{Repulsive Potts Model} Originally introduced in statistical mechanics~\cite{Potts1952}, the $q$-state repulsive Potts model~\cite{Boykov2001} penalizes identical neighboring states, thereby promoting diverse label assignments and encouraging representational differentiation. This is formalized in Eq.~\eqref{eq:repulsive}, where $\beta$ controls the strength of repulsion and $\delta_{rk}$ is the Kronecker delta.
\begin{equation}
    \psi_{ij} (x_r, x_k) = \text{exp}(\beta \cdot \delta_{rk}) \label{eq:repulsive}
\end{equation}
In SAOBP, we adopt this repulsive potential as a pairwise factor function within the belief propagation step to regulate information flow between tokens, promoting attention patterns to diversify rather than localized. 

\section{Self-Attention One-step Belief Propagation}
To address the issue of localization observed in standard self-attention mechanisms, we propose \textbf{SAOBP}, an algorithm designed to explicitly incorporate multi-hop dependencies into the attention updates. Specifically, we interpret the self‐attention score matrix as a factor graph, where each pair of variable nodes is connected via a distinct factor node, as illustrated in Fig.~\ref{fig:saobp-framework}. Instead of performing iterative updates, our experiments suggest that a single-step message passing, when executed alongside the standard parameter update, is sufficient to introduce global contextual information. SAOBP consists of three stages: (1) Initialization, (2) One-step Message Passing, and (3) Final belief update.

\subsection{Notations and Definitions} 
\hspace*{0.7em} \textbf{Node.} Let each row $A_i, A_j \in \mathbb{R}^L$ of the self-attention weight matrix represent nodes $i, j$, where the vector components $(x_{i1}, x_{i2}, \dots, x_{iL})$ denote the attention \textit{label} over all $L$ tokens, respectively. Each components $x_{ik}$ is interpreted as a discrete label reflecting the similarity between token $i$ and indirect token (label) $k$. This representation allows us to apply belief propagation within the self-attention framework by treating each node as a probabilistic variable over token-level assignments. 

\textbf{Messages.} In the factor graph formulation, two types of messages are exchanged: $m_{i \rightarrow f_{ij}}$ (from variable node $i$ to the factor node $f_{ij}$), and $\tilde{m}_{f_{ij} \rightarrow j}$ (from the factor node $f_{ij}$ back to variable node $j$). As illustrated in Fig.~\ref{fig:saobp-framework}, when updating the attention probability for token $A_j$, messages originating from all other tokens $A_m$ $(m \in [1, L], m \neq j)$ pass through their respective pairwise factor nodes before reaching $A_j$. 

\textbf{Pairwise Factor Function.} We adapt Eq.~\eqref{eq:repulsive} as a pairwise factor function to effectively encode global token interactions. The pairwise factor function $\psi_{ij}(x_r, x_k)$ characterizes how indirect tokens dynamically influence the attention score refinement. A tunable parameter $\lambda (\geq 0)$ explicitly controls the strength of repulsive interactions, allowing fine-grained adjustments in attention diversification. We propose \textbf{BP-High} model, which assign higher compatibility scores to dissimilar token (label) pairs ($x_r \neq x_k$). This encourages the attention mechanism to capture a diverse range of global interactions, rather than collapsing onto a limited subset of tokens. 
\begin{equation}
\psi_{ij}(x_r, x_k) =
\begin{cases}
\exp(\lambda), & x_r \ne x_k \\[1.5pt]
1, & x_r = x_k
\end{cases} \label{def:factor-high}
\end{equation}

\subsection{Initialization}
We reuse the result of conventional transformer's trained output weight values as a initializer to preserve normalization and numerical stability. The initializer function (prior) for node $i$ is defined as:
\begin{equation}
\phi_i(x_{k}) = \exp(\log p(x_{ik})) = p(x_{ik}) = A_{ik} \label{eq:initializer}
\end{equation}

\subsection{One step Message Passing}
Building on the standard message-passing rules defined in Eq.\eqref{eq:msg-var2fac}, we derive update equations tailored to the self-attention factor graph in Fig.\ref{fig:saobp-framework}. Since each variable node pair is connected via a single factor node, the product over multiple factors reduces to a single term. Accordingly, the message from factor node $f_{ij}$ to variable node $j$ is computed as shown in Eq.~\eqref{eq:one-step-mp}.
\vspace{-0.10in}
\begin{equation}
\label{eq:one-step-mp}
\small
\begin{split}
    m_{f_{ij}\rightarrow j}^{1}(k) 
    &= \sum_{r=1}^{L}\psi_{ij}(r,k)\phi_i(r) \\[0.1em]
    &= A_{ik} + \exp(\lambda) (\sum_{r=1}^{L}A_{ir} -A_{ik}) \\[0.1em]
    &=A_{ik} + e^{\lambda} (1 - A_{ik})
\end{split}
\end{equation}
The belief estimates of node $j$ after aggregating the incoming messages is approximated as ~\eqref{eq:1step-bp-estimates}. 
\begin{equation}
\label{eq:1step-bp-estimates}
\small
\begin{split}
    b_j^{1}(k) &\propto \prod_{n \in \mathcal{N}(i)} \tilde m^{(t-1)}_{f_{ni} \to i}(k) \\[0.1em]
    &\approx \phi_j(k)\cdot m_{f_{ij}\rightarrow j}^{1}(k) \\[0.1em]
    &= A_{jk}\left[A_{ik}+\exp(\lambda)\left(\sum_{l=1}^{L}A_{il}-A_{ik}\right)\right]
\end{split}
\end{equation}

\subsection{Final Belief Update}
The final belief at node $j$ is computed by collecting all messages from other variable nodes, following the belief update rule in Eq.~\eqref{eq:final‐belief}. The normalization constant $Z_j$ is given by $Z_j = \sum_{x_j} \tilde{b}_j(x_j)$, ensuring the belief distribution forms a valid probability distribution. The final belief assigned to label $k$ for node $j$ is expressed as:
\begin{equation}
\label{eq:1st-final‐belief}
\small
\tilde{b}_j(k) = \frac{A_{jk}}{Z_j} \prod_{i=1}^{L}\left[A_{ik}+\exp(\lambda)\left(\sum_{l=1}^{L}A_{il}-A_{ik}\right)\right]
\end{equation}

\paragraph{Overview} A detailed algorithm of SAOBP is provided in Algorithm~\ref{alg:bp-high}, which follows the corresponding notations introduced in Fig.~\ref{fig:saobp-framework}. The outer loop iterates over each attention row $A_j$ to perform the final belief update, while the inner loop aggregates messages from all other tokens $A_1$ through $A_L$, each connected to $A_j$ via a distinct factor node. The algorithm is implemented using vectorized matrix operations to ensure computational efficiency and scalability.
\begin{algorithm}[h]
  \caption{SAOBP Algorithm (BP-High)}
  \label{alg:bp-high}
  \begin{algorithmic}[1]
    \Require Attention $A \in \mathbb{R}^{L \times L}$, $\lambda$ \hfill \text{\small // Initialization}
    \For{$j = 1$ to $L$}
      \For{$i = 1$ to $L$} \hfill \text{\small // Message passing}
        \State $S_i \gets \sum_{k=1}^{L} A_{ik}$
        \State $M_i \gets e^{\lambda} S_i + (1 - e^{\lambda}) A_i$
        \State $P_i \gets \prod_{k=1}^{L} M_{ik}$
        \State $MP_i \gets P_i / M_i$  \hfill \text{\small // Exclude self-message}
      \EndFor
      \State $UP_j \gets A_j \odot MP$  \hfill \text{\small // Final belief update}
      \State $UP_j \gets UP_j / \sum UP_j$ \hfill \text{\small // Normalization}
    \EndFor
  \end{algorithmic}
\end{algorithm}

\section{Global Token Dependency}
To quantify the extent of multi-hop dependencies---intermediate and indirect interactions typically underrepresented by standard self-attention---we introduce the \textbf{Global Token Dependency (GTD)}. GTD allows us to analyze how effectively models leverage multi-hop relationships across different tasks during inference. It also enables the detection of attention collapse, where distributions concentrate on only a few dominant tokens. Inspired by graph-theoretic diffusion, where information propagation is modeled by summing powers of an adjacency matrix, we reinterpret this concept in the context of language models. Formally, let $A^{(l, h)}$ represent the attention matrix of the $l$-th layer and $h$-th head. Treating $A^{(l, h)}$ as a transition probability matrix, we define cumulative indirect interactions as the discounted sum of attention paths of length $t \geq 2$ between tokens $i$ and $j$:
\begin{equation}
G^{(l,h)}_{ij} = \sum_{t=2}^{K} \beta^{t-1} \left(A^{(l,h)}\right)^t_{ij} \label{eq:glob}
\end{equation}
where $\beta \in (0,1)$ is a discount factor that down-weights longer paths, and $K$ denotes the maximum multi-hop step. Using this \textit{global} matrix $G^{(l,h)}$, we define the GTD as:
\begin{equation}
\mathrm{GTD}(A^{(l, h)}) = \frac{\lVert G^{(l, h)} \rVert_F^2}{\lVert A^{(l, h)} \rVert_F^2 + \lVert G^{(l, h)} \rVert_F^2} \label{eq:gtd}
\end{equation}
where $\lVert \cdot \rVert_F$ denotes the Frobenius norm. Intuitively, GTD quantifies the relative strength of multi-hop interactions compared to the total attention mass. Maintaining GTD within a moderate range is crucial: excessively low values suggest entropy collapse and insufficient global context, whereas excessively high values might indicate noisy or overly diffuse attention patterns detrimental to inference quality.

To further analyze the quality of attention, we introduce the \textbf{Indirect Entropy}, which quantifies the internal uncertainty induced by intermediate token transitions.
\begin{equation}
H_{\mathrm{ind}} = -\frac{1}{L} \sum_{i=1}^L \sum_{j=1}^L \widetilde{G}_{ij} \cdot \log \widetilde{G}_{ij} \label{eq:indirect-entropy}
\end{equation}
$\widetilde{G}_{ij} = \frac{G_{ij}}{\sum_{k=1}^L G_{ik}}$ denotes the normalized probability of transitioning from token $i$ to token $j$ via indirect paths. A low $H_{\mathrm{ind}}$ indicates that multi-hop attention remains deterministic, signaling residual localization. Higher values of $H_{\mathrm{ind}}$ reflect more uniformly distributed global interactions, suggesting that the model integrates diverse contextual information beyond direct attention.

\section{Experimental Settings}
Since our method modifies the self-attention computation mechanism, it is necessary to pretrain models from scratch. We conduct experiments using three BERT-style Transformer architectures that vary in model size: BERT-Mini, BERT-Small, and BERT-Medium (Table~\ref{app_tab:model-info} for specifications). Pretraining is performed on a composite corpus comprising WikiText~\cite{merity2016pointer}, BookCorpus~\cite{zhu2015aligning}, and OpenWebText~\cite{gokaslan2019openweb} to ensure coverage of diverse linguistic patterns. Detailed hyperparameter configurations are provided in Table~\ref{app_tab:pretr_params}. Following pretraining, each model is finetuned on on a diverse set of benchmarks, including GLUE~\cite{wang2018glue}, SQuAD~\cite{rajpurkar2016squad}, HellaSwag~\cite{zellers2019hellaswag}, and RACE-Middle~\cite{lai2017race}. These benchmarks are selected to span a range of contextual demands: from tasks requiring local, short-term understanding (e.g., GLUE) to those necessitating long-range reasoning across broader contexts (e.g., RACE-Middle). This diversity allows us to examine how GTD interacts with model inference across varying levels of contextual complexity. We compute GTD using a discount factor of $\beta = 0.9$ and multi-hop paths up to length $K = 4$.

\paragraph{Model Variants and Baselines}
To examine the role of repulsive Potts-based compatibility of BP-High, we design multiple factor function variants within the SAOBP framework. As a non-Potts baseline, we introduce the \textbf{BP-ElemMul}, which does not employ the repulsion term in Eq.~\eqref{eq:repulsive}. Instead, token compatibility is computed via element-wise inner product between attention rows. The normalization term $Z_i$ ensures stability during training. This formulation allows us to isolate the contribution of the repulsive Potts factor in enhancing attention diversity.
\begin{equation}
A_{ij} = \frac{1}{Z_i} \sum_{k=1}^L A_{ik} * A_{jk}
\label{def:factor-norm}
\end{equation}
\vspace{-0.15em}
To validate the impact of global interaction diversity, we construct \textbf{BP-Low}, which suppresses attention diversity by penalizing dissimilar label pairs with negative weights. This formulation intentionally reduces attention spread by reinforcing token similarity, providing a contrastive counterpart to \textbf{BP-High}.
\begin{equation}
\psi_{ij}(x_r, x_k) =
\begin{cases}
\exp(-\lambda), & x_r \ne x_k \\[1.5pt]
1, & x_r = x_k
\end{cases} \label{def:factor-low}
\end{equation}

To further validate our SAOBP algorithm, we directly implemented two previously proposed methods that aim to address attention entropy collapse, using them as comparative baselines in our study. ~\citet{jha2025entropy} introduce an auxiliary loss function that penalizes low-entropy attention heads (referred to in our experiments as \emph{Entropy-Reg}), while ~\citet{bao2024eigen} propose a strategy to prevent rank collapse by controlling the trace and variance of the query and key weight matrices (referred to as \emph{Eigen-Reg}).

\paragraph{Scaling Repulsive Strength Parameter}
To promote stable training and prevent over-regularization, we scale the repulsive strength hyperparameter $\lambda$ proportionally to model size. For BERT-Mini, we fix $\lambda = 0.2$ due to its relatively small number of layers and attention heads. Based on the ratio of total parameters, we assign $\lambda = 0.08$ for BERT-Small and $\lambda = 0.05$ for BERT-Medium. This scheduling strategy is motivated by prior empirical findings, such as DistilBERT~\cite{Sanh2019DistilBERT} and label smoothing in T5~\cite{Raffel2020Exploring}, which suggest that aggressive regularization may be detrimental in higher-capacity models.

\begin{figure}[t]
\begin{center}
\centerline{\includegraphics[width=\columnwidth]{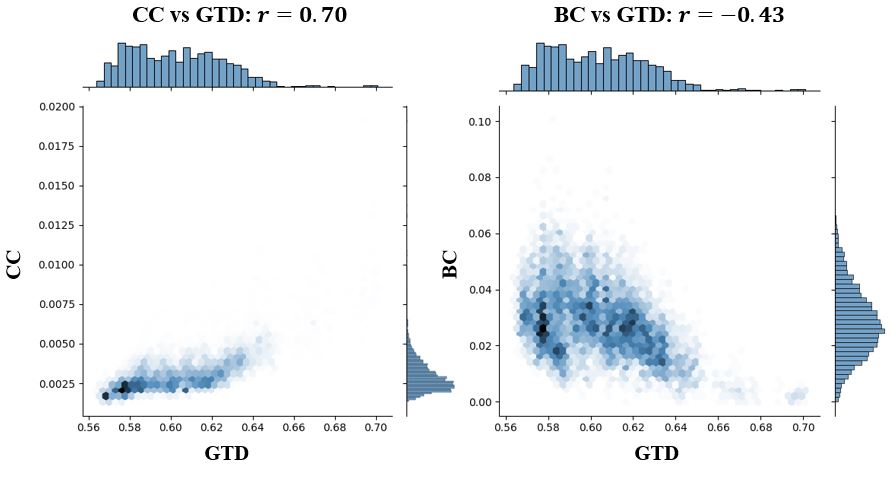}}
\caption{\small Correlation between GTD and graph-theoretic indices CC (left), BC (right) during pretraining. For each sample-head pair in the RACE-Middle dev set, we compute average GTD and project the corresponding attention matrix to a token graph ($\tau = 10^{-4}$) and measure CC, BC on a 40-node subgraph. Histograms on the margins show the empirical density of each variable, with the pearson coefficient $r_{CC}=0.70$, $r_{BC}=-0.43$ under $p<0.01$.}
\label{fig:cc-bc-gtd}
\end{center}
\vspace{-0.35in}
\end{figure}

\section{Results and Analysis}
\begin{figure*}[t]
\begin{center}
\centerline{\includegraphics[width=\linewidth]{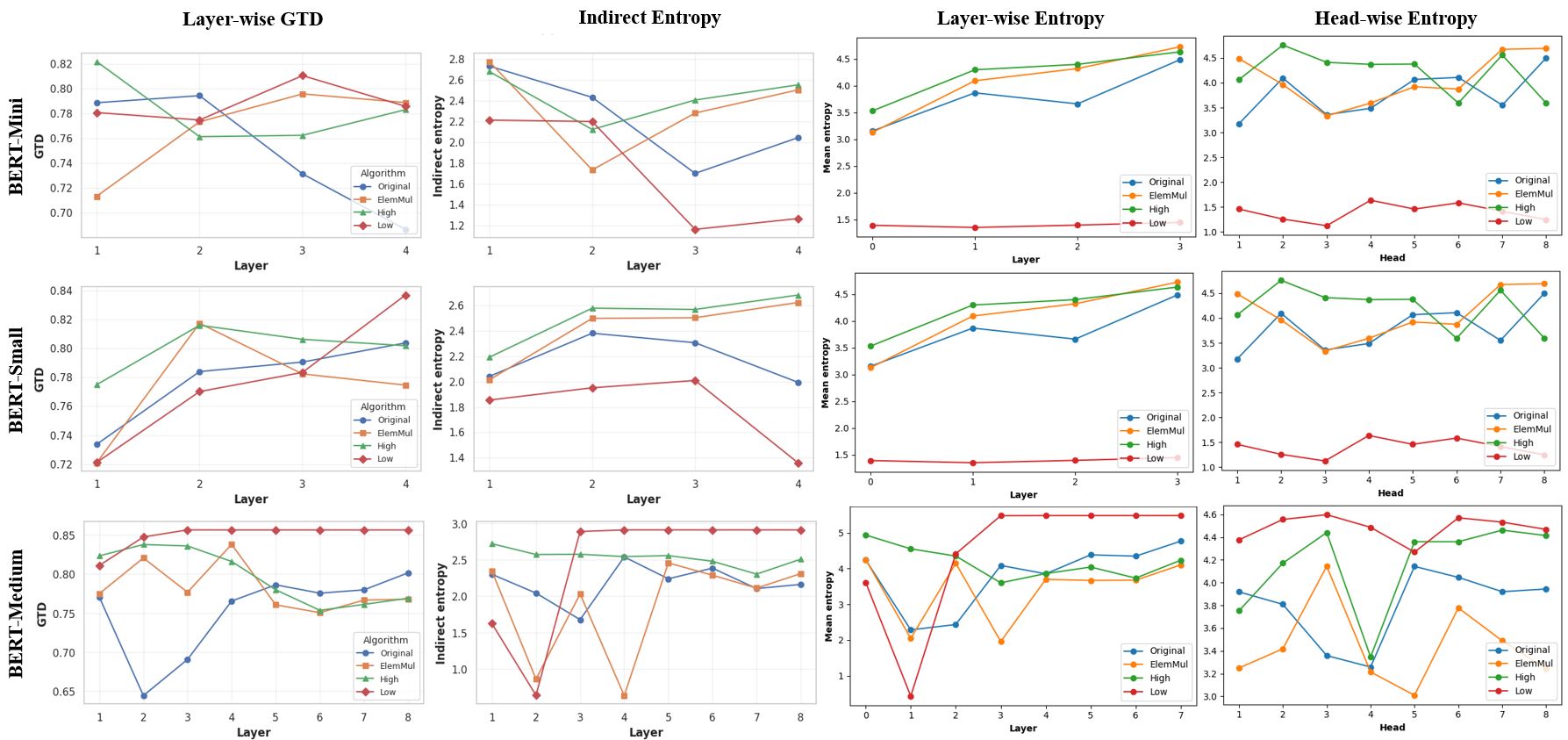}}
\caption{\small Layer-wise GTD and Entropy under Different Factor Functions on 1024 samples of RACE-Middle. Each row corresponds to a model size (top = BERT-Mini, middle = BERT-Small, bottom = BERT-Medium). Columns report, from left to right: (1) GTD ratio per layer, (2) indirect entropy per layer, (3) mean positional entropy across layers, and (4) mean head entropy in the final layer. Curves compare four factor functions—Original (blue), ElemMul (orange), High (green), and Low (red). The x-axis is the Transformer layer (or head index in the last column); the y-axis is the corresponding metric value. Higher GTD and entropy indicate stronger and more evenly distributed multi-hop interactions.}
\label{fig:gtd-analysis}
\end{center}
\vspace{-0.25in}
\end{figure*}

We hypothesize that explicitly modeling multi-hop token dependencies within the self-attention mechanism can alleviate attention localization and improve model performance. To empirically validate this hypothesis, we address the following three research questions. The main analysis is illustrated in Fig.~\ref{fig:gtd-analysis}.

\paragraph{Does GTD effectively capture the locality of attention?}
Before analyzing our proposed algorithm, we first evaluate whether GTD reliably captures indirect, global token interactions and serves as a meaningful indicator correlated with model performance.

Given that few prior studies explicitly quantify intermediate relationships in self-attention, we compare GTD against established graph-theoretic metrics widely utilized in network analysis~\cite{barrat04,freeman77} to assess its reliability. \textit{Clustering Coefficient (CC)} quantifies the mean normalized triangle density in the graph, indicating the extent of local token clustering. Higher CC values implies that tokens are organized into clusters with dense interconnections whereas lower CC corresponds to sparser connectivity~\cite{barrat04}. \textit{Betweenness Centrality (BC)} measures the average unnormalized centrality of nodes based on their presence on shortest paths, highlighting whether attention is concentrated through a small number of intermediate tokens. Lower BC values suggest more evenly distributed information flow~\cite{freeman77}.

Fig.~\ref{fig:cc-bc-gtd} shows that GTD exhibits a positive correlation with CC ($r \approx 0.70$) and a moderate negative correlation with BC ($r \approx -0.43$). This finding aligns with our intended definition of GTD, as higher GTD values indicate globally interconnected attention graphs: multi-hop transitions lead to shorter token-to-token distances (higher CC) and reduce over-reliance on specific bridging tokens (lower BC). However, as shown in Table~\ref{tab:pretrain-gtd}, we also observe cases where CC remains relatively low despite improved accuracy and higher GTD scores. This discrepancy stems from the prevalence of numerous low-weight, long-range edges introduced by SAOBP refinement, which are not fully captured by CC since it focuses solely on dense local clustering. As a result, CC underestimates global connectivity when attention flow becomes more distributed through weaker links, limiting its ability to reflect nuanced token-level interactions. 

Unlike CC and BC, which rely on thresholded or binarized representations of the attention graph and risk losing valuable magnitude information, GTD directly calculates on the fully weighted attention matrix. This approach enables GTD to sensitively detect subtle structural variations in attention flow. Detailed example is reported in Appendix~\ref{sec_app:gtd-analysis}. GTD provides a robust and differentiable measure, reliably capturing nuanced global dependency patterns that significantly impact model dynamics. \emph{These correlations with traditional metrics CC and BC empirically support GTD as an interpretable indicator of global attention distribution.}

\begin{table}[t]
\centering
    \resizebox{\linewidth}{!}{%
    \begin{tabular}{l l cc c cc}
          \toprule
          \multirow{2}{*}{\bfseries Model} & \multirow{2}{*}{\bfseries Algorithm}
           & \multicolumn{2}{c}{\bfseries GTD $\uparrow$}
           & \multirow{2}{*}{\bfseries Acc. $\uparrow$}
           & \multicolumn{2}{c}{\bfseries Graph metrics}\\
          \cmidrule(lr){3-4}\cmidrule(lr){6-7}
           & & GTD & I.E. &  & CC $\downarrow$ & BC $\downarrow$\\
          \midrule
          Mini & Original & 0.75 & 2.20 & 28.48 & 0.25 & 6.82\\
            & High & \textbf{0.78} & \textbf{2.56} & \textbf{29.53} & \textbf{0.16} & \textbf{5.97}\\
            & Low & 0.78 & 1.4 & 23.33 & 0.28 & 8.03\\
            & ElemMul & \textbf{0.77} & \textbf{2.43} & \textbf{29.89} & \textbf{0.24} & \textbf{5.88}\\
          \midrule
          Small & Original & 0.79 & 2.18 & 29.39 & 0.24 & 7.14\\
          & High & \textbf{0.80} & \textbf{2.51} & \textbf{31.34} & \textbf{0.21} & \textbf{6.02}\\
          & Low  & 0.80 & 1.8 & 28.76 & 0.31 & 6.97\\
        & ElemMul & \textbf{0.78} & \textbf{2.41} & \textbf{31.48} & \textbf{0.26} & \textbf{5.48}\\
          \midrule
          Medium & Original & 0.75 & 2.17 & 31.34 & 0.30 & 6.53\\
        & High & \textbf{0.79} & \textbf{2.52} & \textbf{33.77} & \textbf{0.25} & 6.98\\
        & Low  & 0.84 & 2.45 & 25.28 & 0.77 & 2.60\\
        & ElemMul & \textbf{0.78} & 1.87 & \textbf{33.15} & \textbf{0.24} & \textbf{5.83}\\
        \bottomrule
    \end{tabular}
    }
\caption{\small Relationship between GTD, Indirect Entropy (I.E.), accuracy on RACE-Middle, and graph-theoretic metrics (CC and BC) across different algorithms and model sizes. All values are computed as the average over 1,024 samples from the RACE dataset. Models that achieve higher accuracy (bold) generally exhibit elevated GTD and I.E. values, alongside reduced BC scores.}
\label{tab:pretrain-gtd}
\vspace{-0.10in}
\end{table}

\begin{figure}[t]
\begin{center}
\centerline{\includegraphics[width=\columnwidth]{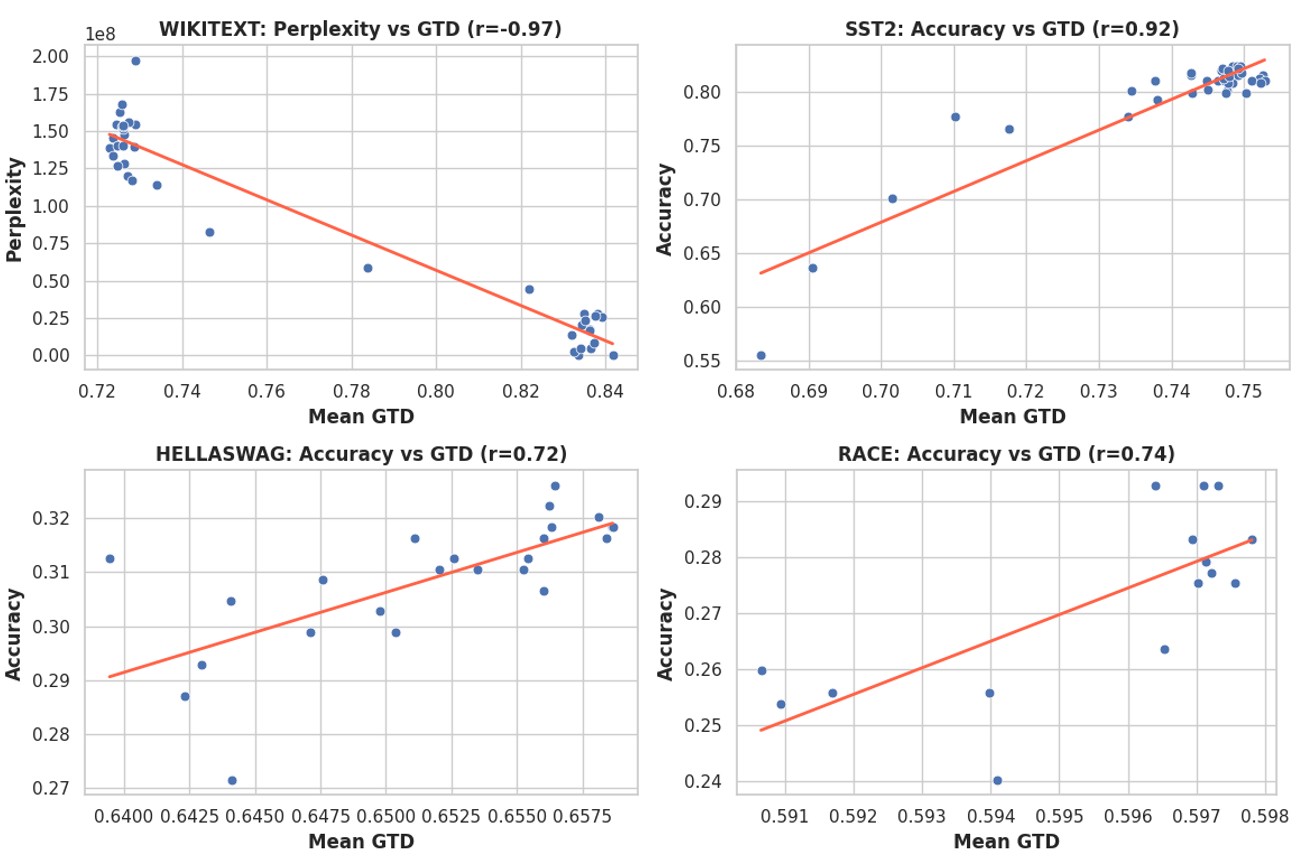}}
\caption{\small Correlation between GTD and model performance in BERT-Mini. We report Pearson correlation coefficients computed over 10–20 checkpoints sampled from the pretraining or finetuning stages ($p < 0.02$). Higher GTD values correlate positively with accuracy on SST2, HellaSwag, RACE tasks, and negatively with perplexity in WikiText.}
\label{fig:gtd-mp-corr}
\end{center}
\vspace{-0.3in}
\end{figure}

To examine GTD's practical utility in interpreting model performance, we analyze its correlations with the model performance across various tasks and checkpoints, as illustrated in Fig.\ref{fig:gtd-mp-corr}. We consistently observe significant correlations ($|r| > 0.5$, $p < 0.02$), showing that \emph{while the extent to which models utilize global contextual information may vary by task, GTD remains a meaningful and interpretable indicator of overall model effectiveness.}

\begin{table}[t]
\centering
\small
\resizebox{\linewidth}{!}{%
\begin{tabular}{llcccc}
\toprule
\bfseries Model & \bfseries Algorithm
    & \bfseries Glue
  & \bfseries Hellaswag
  & \bfseries RACE-Middle
  & \bfseries SQuAD \\
\midrule
\multirow{4}{*}{Mini}
  & Original & 53.37 & 28.06 & 28.48 & 19.22 \\
  & High  & \textbf{54.66} & \textbf{29.55} & 29.53 & \textbf{24.07} \\
  & Low   & 51.89 & 25.59 & 23.33 & 3.99 \\
  & ElemMul  & 54.65 & 29.07 & \textbf{29.89} & 23.03 \\
\midrule
\multirow{4}{*}{Small}
  & Original & 54.03 & 30.30 & 29.39 & 30.08 \\
  & High  & \textbf{57.61} & \textbf{30.79} & 31.34 & 29.65 \\
  & Low   & 51.89 & 26.34 & 28.76 & 14.15 \\
  & ElemMul  & 54.83 & 30.74 & \textbf{31.48} & \textbf{31.23} \\
\midrule
\multirow{4}{*}{Medium}
  & Original & 69.29 & 33.27 & 31.34 & 49.14 \\
  & High  & \textbf{69.89} & \textbf{33.66} & \textbf{33.77} & \textbf{50.35}\\
  & Low   & 40.96 & 26.48 & 25.28 & 0.12 \\
  & ElemMul  & 67.40 & 33.40 & 33.15 & 49.26 \\
\bottomrule
\end{tabular}
}
\caption{\small Accuracy on multiple evaluation tasks.}
\label{tab:extra3}
\vspace{-0.15in}
\end{table}

\paragraph{Does SAOBP mitigate entropy collapse?}
Fig.~\ref{fig:gtd-analysis} presents comprehensive experiments examining the effectiveness of SAOBP algorithms in regulating attention entropy and GTD values. The BP-High variant consistently maintains or enhances both indirect entropy and overall mean entropy, particularly in deeper layers across model sizes. These results confirm that the repulsive Potts factor employed in BP-High effectively counters the entropy collapse commonly observed in Transformer layers. In terms of GTD, BP-High achieves the highest or second-highest values across layers, demonstrating that message passing mechanism effectively incorporates multi-hop contextual information. Overall, we observed that heads with GTD values around $0.6–0.8$ tended to achieve better performance on downstream tasks, whereas heads falling below $\sim$0.5 or above $\sim$0.85 were often associated with degraded performance or unstable training.

Among other SAOBP variants, BP-ElemMul exhibits moderate multi-hop dependency, closely following baseline GTD values in earlier layers but aligning progressively with BP-High in deeper layers, leading to stable entropy levels. This behavior aligns with performance outcomes shown in Table~\ref{tab:extra3}, indicating that similarity-based compatibility functions may be better suited for tasks requiring less aggressive token alignment (e.g., SQuAD, SST2). Conversely, strongly repulsive factors are more effective in facilitating broad, global context reasoning (e.g., RACE-Middle, MNLI). Although BP-Low yields high GTD scores, it significantly reduces (or elevates) entropy, resulting in skewed entropy distributions. This result demonstrates that excessive suppression of repulsive interactions can overly regularize the model and limit representational capacity. More extensive analysis is reported in Appendix~\ref{sec_app:entropy_anlaysis}.

Collectively, these findings substantiate our claim that \emph{BP-High effectively mitigates entropy collapse and modulates GTD within a beneficial range, ensuring stable and expressive attention dynamics across various downstream tasks.} Our results also highlight design trade-offs, motivating future exploration of adaptive factor functions. 

\paragraph{Does SAOBP improve downstream performance?}
As summarized in Table~\ref{tab:extra3}, we compare the original algorithm with SAOBP variants to assess their impact on model performance in downstream tasks. The BP-High consistently achieves enhanced performance along with higher GTD values, suggesting that SAOBP facilitates more globally coherent and evenly distributed attention patterns during inference. Table~\ref{tab:baseline-results} further shows that, compared to other entropy-regulated methods, SAOBP achieves higher accuracy. 
Performance improvements are particularly evident in tasks that require long-range reasoning or cross-sentence inference, such as QNLI and RACE-Middle. Specifically, BP-High applied to BERT-Mini yields accuracy comparable to, or even surpassing, that of the larger BERT-Small model on some benchmarks, including MNLI and QQP. Furthermore, the accuracy gap between the baseline and SAOBP narrow as model size increases, implying that larger models inherently capture global structures through increased depth, while smaller models particularly benefit from explicit multi-hop regularization provided by SAOBP. These findings underscore \emph{the efficacy of SAOBP in enabling compact models to approximate global reasoning capabilities typically observed only in deeper architectures.}

\begin{table}[t]
\centering
\small
\resizebox{\linewidth}{!}{%
\begin{tabular}{lccc}
\toprule
\textbf{Baseline} & \textbf{PPL (WikiText)} & \textbf{QNLI (Acc)} & \textbf{SQuAD (Acc)} \\
\midrule
\textbf{BP-High}     & \textbf{21.86} & \textbf{71.63} & \textbf{29.65} \\
Entropy-Reg          & 26.35 & 70.30 & 26.76 \\
Eigen-Reg            & 23.85 & 71.07 & 26.15 \\
\bottomrule
\end{tabular}
}
\caption{\small Comparison of attention-entropy regularization baselines on WikiText perplexity and accuracy (QNLI, SQuAD).}
\label{tab:baseline-results}
\vspace{-0.15in}
\end{table}

\section{Conclusion}
We propose SAOBP, a novel self-attention refinement framework that leverages structured message passing to explicitly capture intermediate token relationships frequently overlooked by standard attention mechanisms. To rigorously analyze the impact of our proposed approach, we introduce the GTD, a metric designed to quantify the extent of multi-hop information flow and interpretably assess attention patterns in terms of entropy dynamics. Through empirical analyses, we demonstrate that SAOBP effectively alleviates entropy collapse, promotes globally coherent attention distributions, and enhances performance across a diverse range of NLP tasks. Notably, performance improvements from SAOBP are particularly significant in small-scale models, where architectural constraints inherently limit the emergence of global context. Our findings suggest that explicit multi-hop regularization through BP can effectively compensate for reduced depth, enhancing expressivity and generalization in compact Transformers. These insights underscore the broader potential of integrating graph-based inference methodologies with attention architectures to develop lightweight, interpretable, and high-performing small-scale language models.

\section*{Limitations}
While our proposed SAOBP framework shows promising results, several limitations remain. The current SAOBP implementation performs only a single-step belief propagation update. While this approach effectively introduces multi-hop interactions, exploring multi-step message passing may further enhance—or potentially degrade—model representational quality.

We used fixed values for the repulsive strength parameter $\lambda$, but variations could be introduced through layer-wise or head-wise scaling. In addition to repulsive functions as factor functions, other functions that satisfy the property, which is the ability to enhance token interactions, could also be applied as factor functions.

\section*{Acknowledgments}
This work is in part supported by the National Research Foundation of Korea (NRF, RS-2024-00451435(20\%), RS-2024-00413957(20\%)), Institute of Information \& communications Technology Planning \& Evaluation (IITP, RS-2021-II212068(10\%), RS-2025-02305453(15\%), RS-2025-02273157(15\%), RS-2025-25442149(10\%) RS-2021-II211343(10\%)) grant funded by the Ministry of Science and ICT (MSIT), Institute of New Media and Communications(INMAC), and the BK21 FOUR program of the Education, Artificial Intelligence Graduate School Program (Seoul National University), and Research Program for Future ICT Pioneers, Seoul National University in 2025.

\newpage

\bibliography{custom}

\clearpage
\appendix

\section{Experimental Setup}
\label{app:hyper-params}
Table~\ref{app_tab:model-info} and Table~\ref{app_tab:pretr_params} summarize the architectural specifications and training configurations of the models used in our experiments. We report details such as model size, number of training steps, learning rate schedules, and other relevant hyperparameters to ensure reproducibility and fair comparison across model variants. All reported accuracy metrics and plotted results are obtained using a fixed random seed of 42.

\begin{table}[h]
\centering
\resizebox{\columnwidth}{!}{
\begin{tabular}{lccccc}
\toprule
Model & Layers & Hidden & Heads & FFN & \# Params\\
\midrule
BERT-Mini & 4 & 256 & 4 & 1024 & $\sim$11\,M \\
BERT-Small & 4 & 512 & 8 & 2048 & $\sim$29\,M \\
BERT-Medium & 8 & 512 & 8 & 2048 & $\sim$41\,M \\
\bottomrule
\end{tabular}
}
\caption{Transformer structure of models}
\label{app_tab:model-info}
\vspace{-0.15in}
\end{table}

\begin{table}[h]
\centering
\resizebox{\columnwidth}{!}{
\begin{tabular}{lccc}
\toprule
Hyperparameters & BERT-Mini & BERT-Small & BERT-Medium\\
\midrule
max steps & $60000$ &  $150000$  & $245000$\\
learning rate & $5e^{-5}$ & $3e^{-4}$ & $3e^{-4}$ \\
lr scheduler type & cosine & cosine & cosine  \\
warmup steps & $3000$ & $7500$ & $12000$ \\
seed & $42$ & $42$ & $42$ \\
\bottomrule
\end{tabular}
}
\caption{Hyperparameters used during pretraining}
\label{app_tab:pretr_params}
\vspace{-0.15in}
\end{table}

\section{Additional Results of GTD}
\label{sec_app:gtd-analysis}
\begin{figure*}[h]
\begin{center}
\centerline{\includegraphics[width=\textwidth]{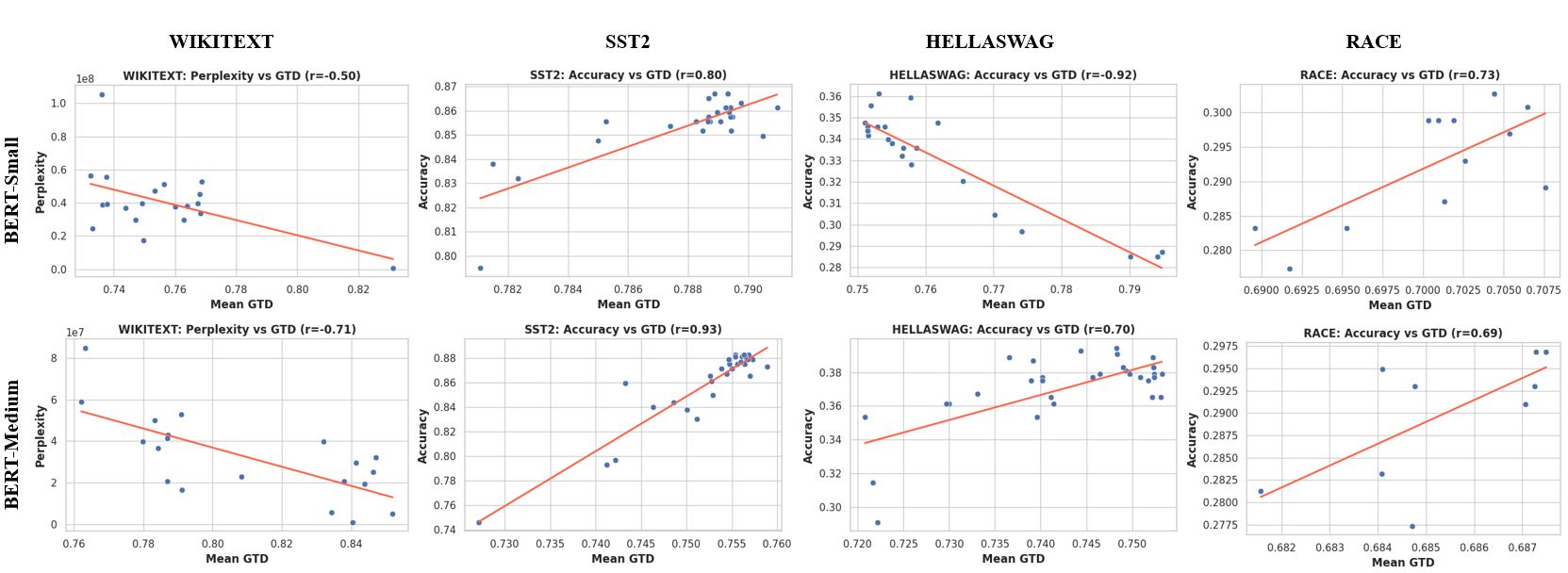}}
\caption{Pearson correlation between GTD and model performance for BERT-Small and BERT-Medium across four benchmarks.  All plotted points represent at least 20 sampled checkpoints per run.}
\label{app_fig:gtd_mp}
\end{center}
\end{figure*}

\paragraph{Correlation Between GTD and Model Performance}
We additionally report Pearson correlation between mean GTD and task performance across additional model sizes to assess the utility of GTD as a diagnostic metric for multi-hop attention interactions. Fig.~\ref{app_fig:gtd_mp} shows that, for every dataset except \textsc{HellaSwag} on the BERT-Small architecture, the absolute correlation exceeds $0.5$ with statistical significance ($p < 0.02$).  The sign reversal observed for \textsc{HellaSwag} on BERT-Small can be attributed to the model’s limited depth: with only four layers, the network is unable to integrate the richer indirect flow captured by high-GTD heads, causing information diffusion that obscures the local stylistic cues critical for this task. These results demonstrate that GTD not only tracks overall performance but also offers insight into architecture–task interactions that govern a model’s internal reasoning dynamics.

\begin{figure}[h]
\begin{center}
\centerline{\includegraphics[width=\columnwidth]{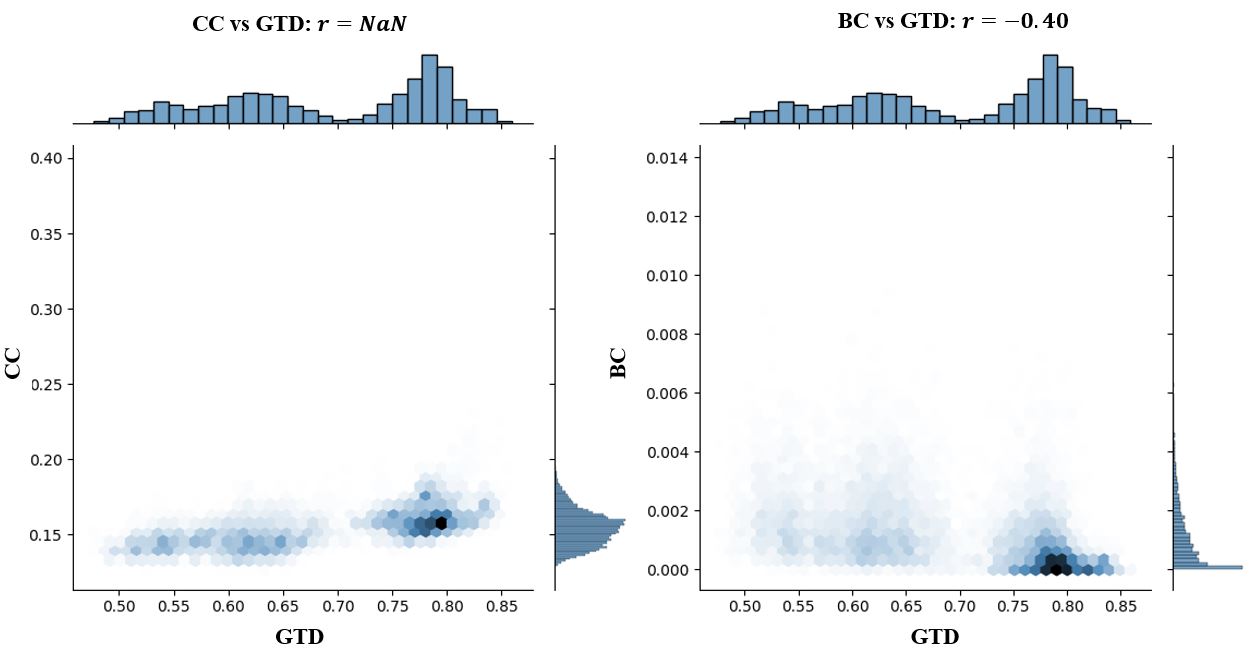}}
\caption{Correlation between GTD and graph-theoretic indices CC (left), BC (right) of RACE-Middle finetuning checkpoints-3116.}
\label{app_fig:cc-bc-gtd}
\end{center}
\vspace{-0.3in}
\end{figure}

\paragraph{Correlation Between GTD and CC/BC metrics}
Fig.~\ref{app_fig:cc-bc-gtd} plots CC and BC against GTD, highlighting their behaviour under varying attention–graph densities. When the attention graph becomes nearly fully connected—--as frequently occurs for tasks that promote widespread token interactions—--the variance of CC collapses, often yielding ill-defined (\texttt{NaN}) values and exposing the metric’s fragility in this regime. BC exhibits a similarly skewed distribution. By contrast, GTD remains well behaved across the full connectivity range, providing a more stable and informative indicator of intermediate, multi-hop dependencies within the attention structure.

\begin{table}[h]
\centering
\small
\resizebox{\linewidth}{!}{%
\begin{tabular}{lcccc}
\toprule
\textbf{GTD} & 
\makecell{\textbf{Wikitext}\\(PPL)} & 
\makecell{\textbf{SST2}\\(Acc)} & 
\makecell{\textbf{Hellaswag}\\(Acc)} & 
\makecell{\textbf{Race-middle}\\(Acc)} \\
\midrule
Single GTD & \textbf{-0.97} & \textbf{0.92} & \textbf{0.72} & \textbf{0.74} \\
Cross GTD  & -0.87 & 0.87 & 0.74 & 0.61 \\
\bottomrule
\end{tabular}
}
\caption{Comparison of single vs cross GTD correlations on Wikitext perplexity and downstream task accuracy (SST2, Hellaswag, Race-middle).}
\label{app_tab:gtd-cross}
\vspace{-0.1in}
\end{table}

\paragraph{Comparison of single versus cross layers GTD}
To better understand the mechanisms through which SAOBP alleviates entropy collapse, we defined token dependency—captured via our proposed GTD measure—at the same unit level as attention entropy, namely within a single attention matrix (i.e., head). Attention entropy is typically computed at the level of a single head, as defined in Eq.~\eqref{eq:attn_ent} of the paper. This allowed us to conduct a parallel analysis based on the information flow in the attention matrix, focusing on token-level interactions at the single-layer, single-head granularity. However, since tokens actually interact within the multi-head Transformer, we additionally compute the Pearson correlation between model performance and cross-layer GTD values (Table~\ref{app_tab:gtd-cross}). All reported correlation coefficients $r$ satisfy $p<0.015$. As our experimental results demonstrate, although token-level interdependencies may be influenced by various components of the Transformer across layers, the single-layer GTD still captures a meaningful aspect of the model’s internal dynamics. It can be effectively used to interpret layer-wise behavior by isolating the influence of each layer’s dynamics on model performance or attention entropy.

\section{SAOBP for Decoder-Only Architectures}
Since decoder-only architectures employ masked self-attention, which differs from the bidirectional attention used in encoder-based models, we made a slight modification to the SAOBP mechanism. Specifically, we blocked message passing from masked (i.e., future) tokens to prevent the propagation of $-\inf$ or noisy values that could interfere with meaningful message computation. Aside from this adjustment, the core logic of SAOBP remains unchanged. 

We pretrained three model sizes (GPT-2 Mini, Small, and Medium) from scratch and evaluated their performance using both perplexity and zero-shot accuracy on downstream tasks. Table~\ref{app_tab:gpt2-results} shows that SAOBP yields strong baselines (with lower perplexity) and suggests that incorporating SAOBP during pretraining can enhance the model’s generalization capabilities, even in decoder-style architectures.

\begin{table}[h]
\centering
\small
\resizebox{\linewidth}{!}{%
\begin{tabular}{llccc}
\toprule
\textbf{Model} & \textbf{Algorithm} & \textbf{Race-middle} & \textbf{BoolQ} & \textbf{PPL (WikiText)} \\
\midrule
\multirow{2}{*}{GPT2-Mini} 
  & Original & 21.44 & 39.54 & 297.1 \\
  & High     & 21.73 & 46.67 & 68.4  \\
\midrule
\multirow{2}{*}{GPT2-Small} 
  & Original & 21.73 & 38.30 & 321.4 \\
  & High     & 22.50 & 39.60 & 75.0  \\
\midrule
\multirow{2}{*}{GPT2-Medium} 
  & Original & 20.04 & 39.60 & 67.5  \\
  & High     & 21.10 & 39.57 & 6.8   \\
\bottomrule
\end{tabular}
}
\caption{\small Comparison of decoder-only models using SAOBP and the original baseline on Race-middle, BoolQ, and WikiText perplexity.}
\label{app_tab:gpt2-results}
\vspace{-0.1in}
\end{table}

\begin{figure*}[t]
\begin{center}
\centerline{\includegraphics[width=\textwidth]{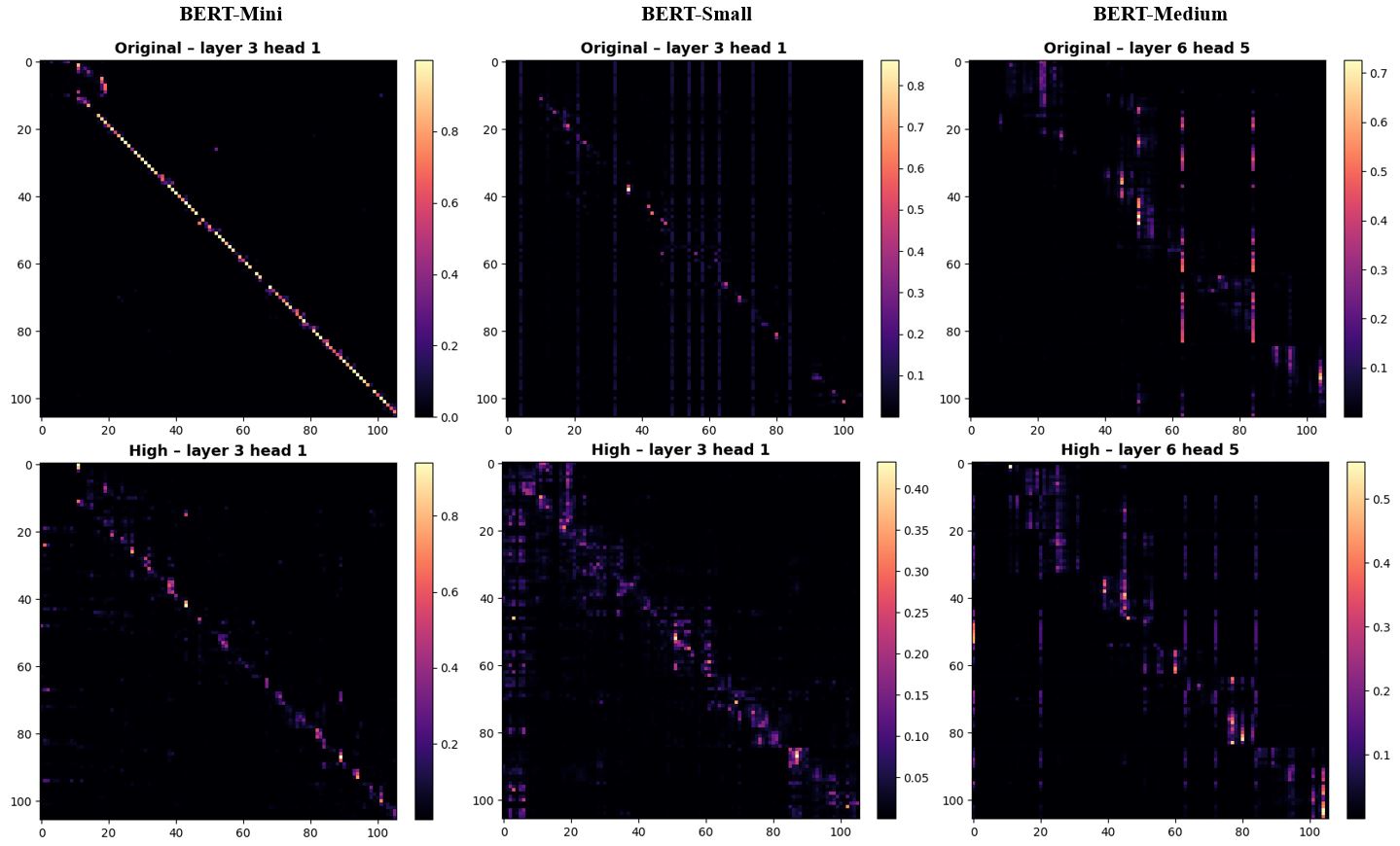}}
\caption{Comparison of attention map sparsity between the original model (top) and BP-High (bottom). Attention maps are visualized using input examples from the SQuAD dataset with a maximum token length of 120. The results indicate that the BP-High algorithm consistently reduces attention sparsity, suggesting enhanced representational capacity through denser attention patterns.}
\label{app_fig:sparse-attn}
\end{center}
\end{figure*}

\section{Detailed Model Performance on Glue}
\label{sec_app:glue_details}
Detailed results for the individual GLUE benchmarks are presented in Table~\ref{tab:glue}.  While the best-performing variant differs slightly by task, BP-High delivers consistent, moderate improvements across the majority of benchmarks.

\begin{table*}[!ht]
\centering
\resizebox{\textwidth}{!}{ 
\begin{tabular}{llcccccccccc}
\toprule
\bfseries Model & \bfseries Algorithm
  & \bfseries CoLA
  & \bfseries MRPC
  & \bfseries RTE
  & \bfseries SST2
  & \bfseries STS-B
  & \bfseries WNLI
  & \bfseries QNLI
  & \bfseries MNLI
  & \bfseries QQP
  & \bfseries Avg. \\
\midrule
\multirow{4}{*}{Mini}
  & Original & 15.95 & 65.93 & 51.99 & 81.88 & 18.20 & 46.48 & 61.50 & 60.14 & 78.31 & 53.37 \\
  & High  & 13.57 & 65.93 & 53.07 & 79.70 & 20.27 & 45.07 & 61.52 & 69.99 & 82.79 & 54.66 \\
  & Low   &  8.17 & 65.44 & 47.65 & 79.93 & 18.41 & 47.89 & 63.04 & 62.55 & 73.96 & 51.89 \\
  & ElemMul  & 12.02 & 65.93 & 55.23 & 79.24 & 20.06 & 43.66 & 63.08 & 69.96 & 82.68 & 54.65 \\
\midrule
\multirow{4}{*}{Small}
  & Original & 28.63 & 68.14 & 50.90 & 87.04 & 23.43 & 23.94 & 71.68 & 70.73 & 85.18 & 56.63 \\
  & High  & 32.38 & 69.12 & 54.51 & 86.35 & 23.90 & 26.76 & 71.63 & 72.43 & 84.81 & 57.99 \\
  & Low   & 29.65 & 69.12 & 53.07 & 85.44 & 22.28 & 30.99 & 71.59 & 69.47 & 85.03 & 57.52 \\
  & ElemMul  & 31.40 & 69.12 & 51.26 & 85.67 & 22.83 & 23.94 & 72.05 & 70.78 & 84.94 & 56.89 \\
\midrule
\multirow{4}{*}{Medium}
  & Original & 28.77 & 75.49 & 53.79 & 87.27 & 78.13 & 53.52 & 82.46 & 75.65 & 88.24 & 69.29 \\
  & High  & 29.98 & 75.98 & 56.32 & 87.39 & 80.84 & 50.70 & 83.80 & 75.84 & 88.13 & 69.89 \\
  & Low   &  0 & 68.38 & 47.29& 50.92 & 0.3 & 56.34 & 49.46 & 32.74 & 63.18 & 40.96 \\
  & ElemMul  & 25.87 & 76.47 & 53.07 & 86.58 & 75.28 & 45.07 & 82.50 & 74.48 & 87.29 & 67.40 \\
\bottomrule
\end{tabular}
}
\caption{BERT-Mini, BERT-Small, Bert-Medium Accuracy on GLUE}
\label{tab:glue}
\end{table*}

\section{Attention Sparsity Profile}
\label{sec_app:sparsity}
Excessive localization in self-attention can induce attention collapse, which is indirectly observable through increased sparsity in the attention maps. To investigate this, we visualize representative attention distributions from deeper layers of each model size, as illustrated in Fig.~\ref{app_fig:sparse-attn}. In the BERT-Mini model, the original attention exhibits a mean sparsity of 0.906, whereas the SAOBP-High variant reduces this to 0.895. A similar trend is observed in BERT-Small (0.891 → 0.883) and BERT-Medium (0.918 → 0.868), suggesting that the proposed algorithm consistently promotes denser and potentially more expressive attention patterns across different model scales.

\section{Attention Entropy Profile}
\label{sec_app:entropy_anlaysis}
\begin{figure*}[h]
\begin{center}
\centerline{\includegraphics[width=\textwidth]{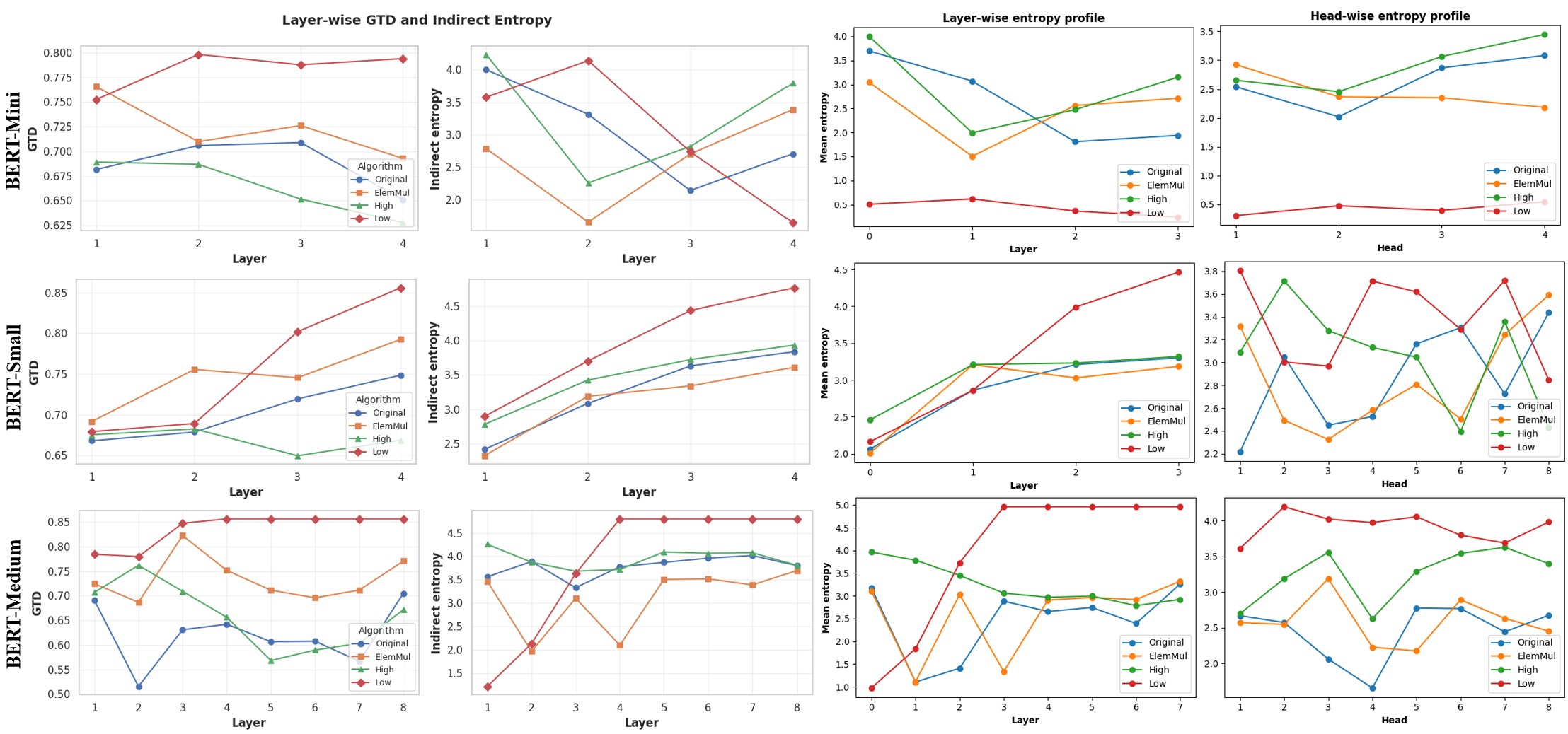}}
\caption{\small GTD value, indirect entropy, layer-wise and head-wise entropy distriutions on SQuAD datasets.}
\label{app_fig:ent_ana_sqd}
\end{center}
\end{figure*}

\begin{figure*}[h]
\begin{center}
\centerline{\includegraphics[width=\textwidth]{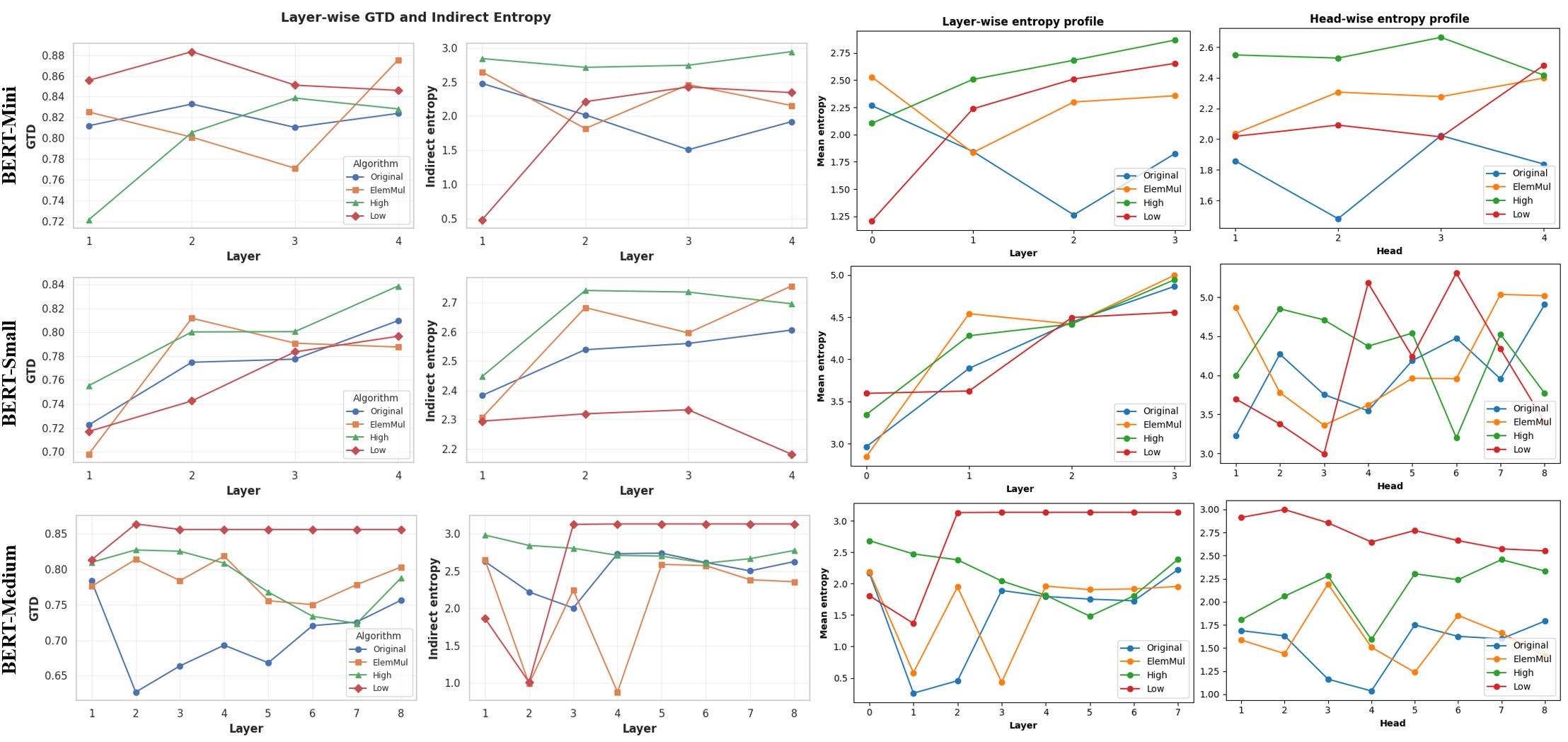}}
\caption{\small GTD value, indirect entropy, layer-wise and head-wise entropy distriutions on SST2 (GLUE) datasets.}
\label{app_fig:ent_ana_sst}
\end{center}
\end{figure*}

We extend the entropy analysis to additional SQuAD (Fig.~\ref{app_fig:ent_ana_sqd}) and GLUE (Fig.~\ref{app_fig:ent_ana_sst}) tasks. Across both corpora, SAOBP consistently alleviates the entropy collapse phenomenon that is most pronounced in smaller models, while preserving high head-level entropy in larger configurations. Additionally, the optimal GTD level appears to be task-dependent: for SQuAD, the best-performing variant exhibits a lower GTD, whereas for SST2 a higher GTD correlates with superior accuracy. By jointly examining GTD, indirect entropy and layer/head-wise total entropy, we can trace how individual layers and attention heads modulate multi-hop information flow, offering a finer-grained view of each model’s internal reasoning strategy.

\section{Computational Cost of SAOBP}
\label{app_tab:cost}
As SAOBP performs an additional refinement over the original self-attention values, it incurs a slight increase in per-step runtime (Table~\ref{app_tab:computation-cost}). However, it achieves faster loss convergence than the baseline, particularly in smaller models.

\begin{table*}[t]
\centering
\resizebox{\linewidth}{!}{
\begin{tabular}{l l c c c c c c c}
\toprule
\textbf{Model} & \textbf{} & 
\makecell{\textbf{step\_per\_s}\\(ms)} & 
\makecell{\textbf{tok\_per\_s}\\(tok/s)} & 
\makecell{\textbf{FLOPs}\\(GFLOPs)} & 
\textbf{Inference} & 
\makecell{\textbf{Latency/batch}\\(ms)} & 
\makecell{\textbf{Throughput}\\samples/s (Mtok/s)} & 
\makecell{\textbf{FIOPs}\\(GFLOPs)} \\
\midrule
Mini & Original & 28.03 & 4567.34 & 1.41 & Original & 7.4  & 4353.7 (0.56) & 0.40 \\
     & High     & 37.40 & 3422.20 & 1.45 & High     & 10.4 & 3085.8 (0.39) & 0.44 \\
     & Low      & 48.54 & 2637.02 & 1.45 & Low      & 9.8  & 3260.0 (0.42) & 0.44 \\
     & Elemul   & 52.88 & 2420.48 & 1.48 & Elemul   & 10.8 & 2963.2 (0.38) & 0.47 \\
\midrule
Small& Original & 42.15 & 3036.74 & 3.65 & Original & 8.7  & 3679.7 (0.47) & 1.61 \\
     & High     & 59.64 & 2146.32 & 3.72 & High     & 14.3 & 2241.3 (0.29) & 1.68 \\
     & Low      & 45.06 & 2840.56 & 3.72 & Low      & 14.1 & 2272.6 (0.29) & 1.68 \\
     & Elemul   & 48.30 & 2649.87 & 3.78 & Elemul   & 11.6 & 2752.0 (0.35) & 1.75 \\
\midrule
Medium& Original& 55.41 & 2309.89 & 5.26 & Original & 17.2 & 1859.5 (0.24) & 3.23 \\
      & High    & 63.40 & 2018.91 & 5.40 & High     & 24.5 & 1308.3 (0.17) & 3.36 \\
      & Low     & 58.75 & 2178.76 & 5.40 & Low      & 24.3 & 1314.3 (0.17) & 3.36 \\
      & Elemul  & 61.40 & 2084.54 & 5.53 & Elemul   & 23.2 & 1376.4 (0.18) & 3.56 \\
\bottomrule
\end{tabular}
}
\caption{The average computational cost over 200 steps during pretraining and inference.}
\label{app_tab:computation-cost}
\end{table*}

\end{document}